\documentclass[letterpaper, 10 pt, conference]{ieeeconf}  

\IEEEoverridecommandlockouts                              
\usepackage{graphics} 
\usepackage{epsfig} 
\usepackage{times} 
\usepackage{amsmath} 
\usepackage{amssymb}  
\usepackage{xcolor}
\usepackage{hyperref}
\usepackage{threeparttable}
\usepackage{multirow}
\usepackage{booktabs}
\usepackage{easyReview}

\title{\LARGE \bf
	BiRP: Learning Robot Generalized Bimanual Coordination using\\ Relative Parameterization Method on Human Demonstration 
}

\author{Junjia Liu, Hengyi Sim, Chenzui Li, and Fei Chen$^{\dagger}$, \IEEEmembership{Senior Member, IEEE}
	\thanks{This work was supported in part by the Research Grants Council of the Hong Kong SAR under Grant 24209021, 14222722 and C7100-22GF and in part by CUHK Direct Grant for Research under Grant 4055140.}
	\thanks{Junjia Liu, Hengyi Sim, Chenzui Li, and Fei Chen are with the Department of Mechanical and Automation Engineering, T-Stone Robotics Institute, The Chinese University of Hong Kong, Hong Kong SAR (e-mail: {\tt\small jjliu@mae.cuhk.edu.hk, hysim@mae.cuhk.edu.hk, czli@mae.cuhk.edu.hk, f.chen@ieee.org}).}
\thanks{$^\dagger$Corresponding authors}
}

\begin{document}

\maketitle
\thispagestyle{empty}
\pagestyle{empty}

\begin{abstract}
	Human bimanual manipulation can perform more complex tasks than a simple combination of two single arms, which is credited to the spatio-temporal coordination between the arms. However, the description of bimanual coordination is still an open topic in robotics. This makes it difficult to give an explainable coordination paradigm, let alone applied to robotics. In this work, we divide the main bimanual tasks in human daily activities into two types: \textit{leader-follower} and \textit{synergistic} coordination. Then we propose a relative parameterization method to learn these types of coordination from human demonstration. It represents coordination as Gaussian mixture models from bimanual demonstration to describe the change in the importance of coordination throughout the motions by probability. The learned coordinated representation can be generalized to new task parameters while ensuring spatio-temporal coordination. We demonstrate the method using synthetic motions and human demonstration data and deploy it to a humanoid robot to perform a generalized bimanual coordination motion. We believe that this easy-to-use bimanual learning from demonstration (LfD) method has the potential to be used as a data augmentation plugin for robot large manipulation model training. The corresponding codes are open-sourced in \url{https://github.com/Skylark0924/Rofunc}.
	
\end{abstract}

\section{Introduction}
Humanoid robots with high redundancy are expected to perform complex manipulation tasks with human-like behavior. However, ensuring the coordination between multiple degrees of freedom is still an open problem in robotics. This is often the key to the success of most human daily activities, like stir-frying, pouring water, sweeping the floor, and putting away clothes. Thus, it is necessary to provide an explainable paradigm to describe and learn coordination. Learning the manipulation of a humanoid robot by observing human motion and behavior is a straightforward idea \cite{yao2021hand}, but the technology behind it is still challenging. It needs to understand human motion data and design a bridge connecting humans and robots. In this work, we focus on the learning and generalization of bimanual coordination motions from human demonstration.

\begin{figure}[ht]
	\centerline{\includegraphics[width=0.9\linewidth]{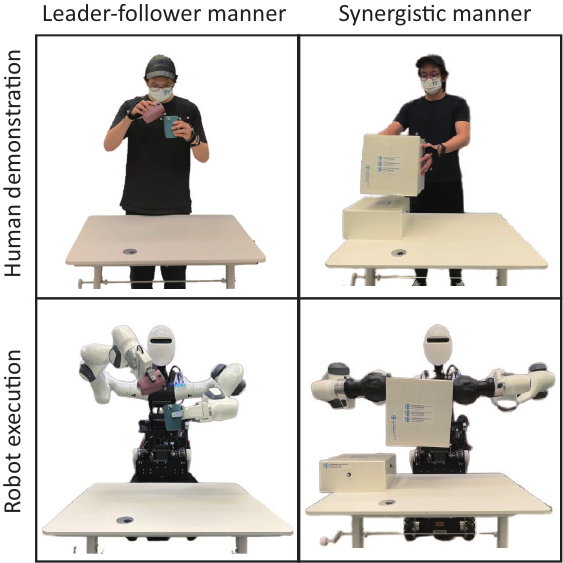}} 
	\caption{The two main bimanual coordination manners in human daily activities are \textit{leader-follower} and \textit{synergistic} coordination. Learning these coordination manners from human demonstration requires the ability to extract the implicit coordination information from motion data and deploy it into new situations with different task parameters.}
	\label{head}
\end{figure}
Learning from demonstration (LfD) is a type of machine-learning approach that allows robots to learn tasks or skills from human demonstrations. Instead of programming robot motions with explicit instructions that are defined manually for each task \cite{lee2015redundancy}\cite{shi2017robust}, LfD enables robots to learn skills by observing human performance \cite{argall2009survey}. It is implemented by the following processes: recording human demonstration data, learning the representation of multiple demonstrations, transferring the data to the workspace of robots, and finally designing a controller for generating the smooth trajectory and its corresponding control commands. LfD has become an increasingly popular approach for training robots, as it can be faster and more efficient than traditional programming methods. It also allows robots to learn tasks that may be difficult to program explicitly, such as those that involve complex movements or interactions with a dynamic environment. Meanwhile, another important feature of LfD is that it enables robots to adapt to new or changing environments \cite{pastor2009learning}, as they can learn from demonstrations in different settings and apply that knowledge to new situations. 

\begin{figure*}[ht]
	\centerline{\includegraphics[width=1\linewidth]{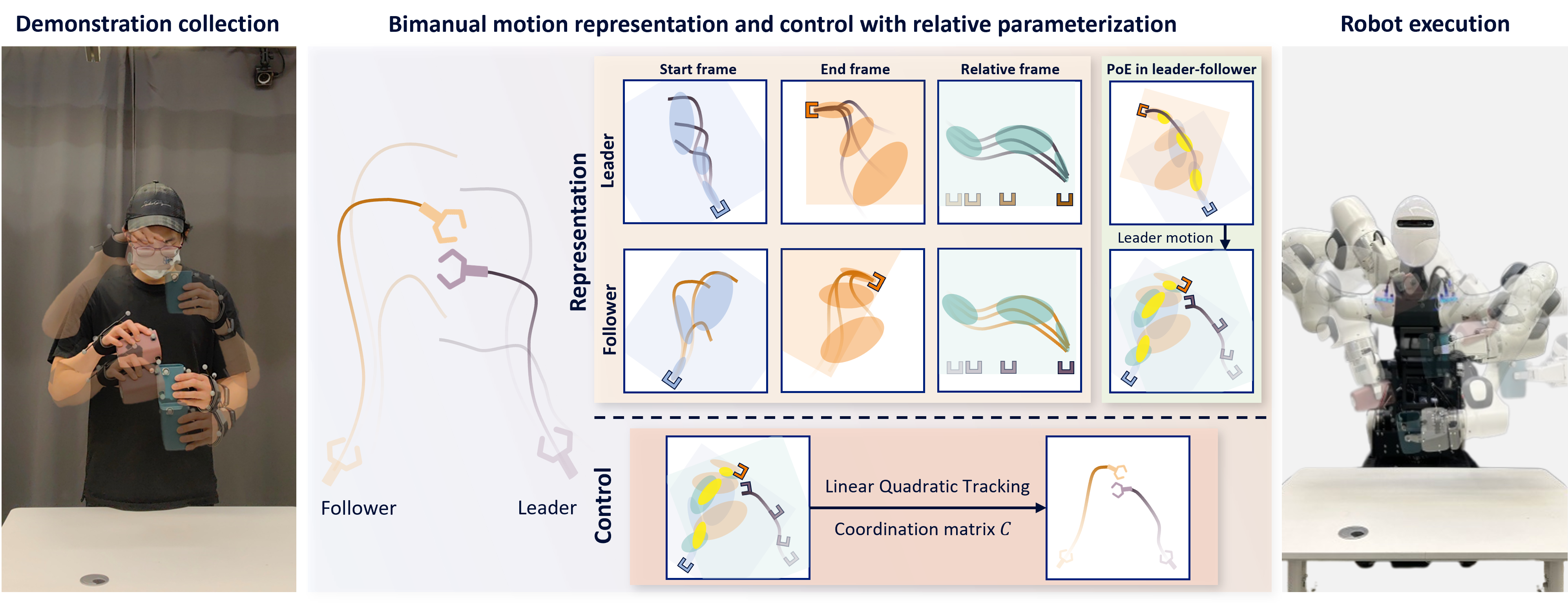}} 
	\caption{The whole framework is illustrated based on a pouring task, which is a specific \textit{leader-follower} example. It starts from the collection process, where the coordinated motion of human arms and the displacement of objects were recorded by Optitrack. Then the coordinated motions were represented by TP-GMM with relative parameterization, and the task-specific GMM was reconstructed in a \textit{leader-follower} manner (given a pre-generated leader motion to construct the corresponding follower motion representation). Follower motion can be generated by LQT (with coordination matrix) for robot execution. }
	\label{framework}
\end{figure*}

Bimanual robots are much more complex to learn from demonstration than single-armed robots that can be taught by kinesthetic teaching \cite{ureche2018constraints}. Some previous works tried to combine the trajectories taught multiple times to realize the kinesthetic teaching of highly redundant robots \cite{gribovskaya2008combining}. However, this also makes the demonstration data less reliable. Recently, several works proposed feasible frameworks for learning directly from human demonstration. Krebs et al. provided a taxonomy of human bimanual manipulation in daily activities by focusing on different types of coordination \cite{krebs2022bimanual}. Liu et al. regarded the leader-follower coordination as sequence transduction and designed a coordination mechanism based on Transformer model to achieve a human-level stir-fry task \cite{liu2022robot}. Besides, offline reinforcement learning algorithms have been used to let robots bimanual coordination tasks from offline demonstration dataset \cite{sun2022mixline}, allowing the robot to learn the most efficient and effective ways to coordinate its arms for a given task.

In this work, we aim to propose an explainable paradigm for learning generalized coordination from demonstration. The main contributions can be summarized as follows:
\begin{itemize}
	\item \textbf{Coordination parameterization:} We propose a relative parameterization method (BiRP) for extracting the coordination relationship from human demonstration and embedding it into the motion generation of each arm.
	\item \textbf{Leader-follower motion generation:} We provide conditional coordinated motion generation for bimanual tasks with different roles in arms, allowing us to generate the follower's motion according to the leader.
	\item \textbf{Synergistic motion generation:} For tasks where there is no obvious role difference between arms, we also provide a motion generation method that enables both arms to adapt to new situations synergistically.
	
\end{itemize}

\section{Construct Bimanual Coordination by Relative Parameterization} 
The definition of relative parameterization is a way to parameterize the relative relationship between bimanual arms and embed this relationship into the representation of each arm. The relative relationship can have many forms, which depend on task-specific coordination characteristics. For example, if bimanual arms are asked to grasp the same object simultaneously and keep the hold until it is placed, the relative relationship can be the relative displacement of end-effectors. The definitions of symbols are listed in Table \ref{term}. 

\begin{table}[t]
	\centering
	\caption{Definition of Symbols}
	\label{term}
	\begin{threeparttable}
		\begin{tabular}{ll}
			\toprule 
			Symbol  &  Definition \\
			\midrule
			$D$ & State dimensions \\
			$O$ & Order of the controller \\
			$P$ & Number of reference frames \\
			$H$ & Number of arms, $h$ refers to left or right arm here \\
			$T$ & Time horizon, $t\in [0, T]$ \\
			$K$ & Number of Gaussian components in a mixture model \\
			$\boldsymbol{\xi}$ & Demonstration motion, $\boldsymbol{\xi}\in  \mathbb{R}^{DT}$  \\
			$\boldsymbol{\zeta}_j$ & Motion in frame $j$,  $\boldsymbol{\zeta}=\left[\zeta_1^\top, \ldots, \zeta_T^\top\right] \in \mathbb{R}^{DOT}$\\
			$\boldsymbol{u}$ & Control command, $\boldsymbol{u}=\left[u_1^\top, \ldots, u_{T-1}^\top\right] \in \mathbb{R}^{D (T-1)}$\\
			$\boldsymbol{x}$ & Robot motion, $\boldsymbol{x}=\left[x_1^\top, \ldots, x_T^\top\right] \in \mathbb{R}^{DT}$\\
			$\boldsymbol{Q}$ & The required tracking precision matrix\\
			$\boldsymbol{R}$ & The cost matrix on control command\\
			\bottomrule[1pt]
		\end{tabular}
	\end{threeparttable}
\end{table}

In this section, we first briefly introduce the fundamental learning from demonstration method used in uni-manual scenarios (Sec. \ref{IIA}), which consists of two parts: demonstration representation and motion reproduction or generation. We add the concept of relative parameterization to these two parts so that both the process of representation learning (Sec. \ref{IIB}) and the process of control (Sec. \ref{IIC}) take into account the bimanual coordination characteristics in the demonstration data. These two methods can be used independently or jointly. A feasible weighting approach is also proposed to increase the importance of coordination characteristics in the representation and control (Sec. \ref{IID}). The whole framework that is illustrated by a \textit{leader-follower} example is shown in Fig. \ref{framework}.

\subsection{Demonstration Representation and Motion Generation} \label{IIA}
Learning from demonstration method is a bridge between humans and robots, which is required to have the ability to extract the characteristics of human skills, plan the trajectories, and control the robot to perform similar skills. Thus, it is necessary to combine human skill learning and robot motion planning and control together in the same encoding approach. A popular way is to link them in the form of probability, like Hidden Markov Model (HMM) and Gaussian Mixture Model (GMM). Besides, considering that the application scenarios of service robots are unstructured and need to adapt to changing situations, a class of task-parameterized models is proposed to address this problem \cite{calinon2013improving}. The \textit{task parameters} are variables describing the task-specific situation, like the position of an object in a pick-and-place task. By contrary, some task-independent information can also be extracted from the demonstration data, which reflects the nature of the skill itself, namely \textit{skill parameters}. The concept of task-parameterized models is to observe the skill in multiple frames, like from starting points and ending points, and describe the impedance of the systems by variations and correlations with a linear quadratic regulator, which can then be used to control the robot.

Task-parameterized Gaussian Mixture Model (TP-GMM) is a typical method that probabilistically encodes datapoints, and the relevance of candidate frames $P$ by mixture models, which has good generalization capability \cite{calinon2016tutorial}. Formally, if we define the task parameters as $\left\{\boldsymbol{b}_j, \boldsymbol{A}_j\right\}_{j=1}^P$, the demonstrations $\boldsymbol{\xi}$ can be observed as $\boldsymbol{\zeta}_j=\boldsymbol{A}_j^{-1}\left(\boldsymbol{\xi}-\boldsymbol{b}_j\right)$ in each frame $j$. These transformed demonstrations are then represented as GMM $\{\pi^{(k)},\{\boldsymbol{\mu}_j^{(k)},  \boldsymbol{\Sigma}_j^{(k)}\}_{j=1}^P\}_{k=1}^K$ by log-likehood maximization, where  $\pi^{(k)}$ refers to prior probability of $k$-th Gaussian component, $\boldsymbol{\mu}_j^{(k)}$ and $\boldsymbol{\Sigma}_j^{(k)}$ refer to mean and covariance matrix of the $k$-th Gaussian in frame $j$. We can regard these Gaussian components in multiple frames as skill parameters that can be transferred following the change of task parameters. For instance, if a new situation is given by task parameters $\{\boldsymbol{\hat{b}}_{j}, \boldsymbol{\hat{A}}_{j}\}_{j=1}^P$, a new task-specific GMM can be generated by Product of Expert (PoE):
\begin{equation}
	\begin{small}
		\begin{aligned}
			\mathcal{N}\left(\boldsymbol{\hat{\nu}}^{(k)}, \boldsymbol{\hat{\Gamma}}^{(k)}\right) \propto \prod_{j=1}^P \mathcal{N}\left(\boldsymbol{\nu}_{j}^{(k)}, \boldsymbol{\Gamma}_{j}^{(k)}\right)
		\end{aligned}
	\end{small}
\end{equation}
where $\boldsymbol{\nu}_{j}^{(k)}=\boldsymbol{A}_j\boldsymbol{\mu}_{j}^{(k)}+\boldsymbol{b}_j, \boldsymbol{\Gamma}_{j}^{(k)}=\boldsymbol{A}_j\boldsymbol{\Sigma}_{j}^{(k)}\boldsymbol{A}_j^\top$.The result of the Gaussian product is given analytically by
\begin{equation}
	\begin{small}
		\begin{aligned}
			\boldsymbol{\hat{\Gamma}}^{(k)}=\left(\sum_{j=1}^P \boldsymbol{\hat{\Gamma}}_{j}^{(k)}{ }^{-1}\right)^{-1}, \quad \boldsymbol{\hat{\nu}}^{(k)}=\boldsymbol{\hat{\Gamma}}^{(k)} \sum_{j=1}^P \boldsymbol{\Gamma}_{j}^{(k)}{ }^{-1} \boldsymbol{\nu}_{j}^{(k)}
		\end{aligned}
	\end{small}
\end{equation}

For generating robot motion from GMM, optimal control methods like Linear Quadratic Regulator (LQR) and Linear Quadratic Tracking (LQT) can be used as planning and control methods. Here we give the classical form of LQT as follows:
\begin{equation}
	\begin{small}
		\begin{aligned}
			cost=\left(\boldsymbol{\hat\nu}-\boldsymbol{x}\right)^\top \boldsymbol{Q}\left(\boldsymbol{\hat\nu}-\boldsymbol{x}\right)  + \boldsymbol{u}^\top \boldsymbol{R} \boldsymbol{u}
		\end{aligned}\label{LQT}
	\end{small}
\end{equation}
where $\boldsymbol{\hat\nu}$ is the mean matrices of the task-specific GMM obtained by the previous PoE process.

Assume that the system evolution is linear, 
\begin{equation}
	\begin{small}
		\begin{aligned}
			x_{t+1}=\boldsymbol{A}_{s} x_t+\boldsymbol{B}_{s} u_t
		\end{aligned}\label{linear_system}
	\end{small}
\end{equation}
where $\boldsymbol{A}_s, \boldsymbol{B}_s$ are coefficients for this system. Then, the relationship between the control command and the robot states can be described in the matrix as $\boldsymbol{x}=\boldsymbol{S}_{x} x_1+\boldsymbol{S}_{u} \boldsymbol{u}$, where $\boldsymbol{S}_{x}\in \mathbb{R}^{DT\times D}$ and $\boldsymbol{S}_{u}\in \mathbb{R}^{DT\times D(T-1)}$ are the matrix form combination of $\boldsymbol{A}_s, \boldsymbol{B}_s$. More details can be found in the appendix of \cite{calinon2016tutorial}. 

Here we just consider an open loop controller, which solution can be given analytically by 
\begin{equation}
	\begin{small}
		\hat{\boldsymbol{u}}=\left(\boldsymbol{S}_{u}^{\top} \boldsymbol{Q} \boldsymbol{S}_{u}+\boldsymbol{R}\right)^{-1} \boldsymbol{S}_{u}^{\top} \boldsymbol{Q}\left(\boldsymbol{\hat\nu}-\boldsymbol{S}_{x} x_1\right)
	\end{small}
\end{equation}
with a residual as $\hat{\boldsymbol{\Sigma}}_{u}=\left(\boldsymbol{S}_{u}^{\top} \boldsymbol{Q} \boldsymbol{S}_{u}+\boldsymbol{R}\right)^{-1}$.

\subsection{Representation with Relative Parameterization}\label{IIB}
In the bimanual setting, coordination is reflected at the data level as some characteristics of the relative motion of the arms. For instance, for a bimanual box-lifting task, this characteristic manifests itself as the arms move from free movement to a fixed relative relationship and maintain this relationship for a certain time frame. For a \textit{leader-follower} task like stir-fry \cite{liu2022robot}, the characteristic refers to the following arm (holding the spoon) motion, and its periodicity is determined with reference to the leading arm (holding the pot). In this work, instead of pre-defining the roles between the arms (as leader or follower), we aim to describe the relative relationship between the arms in a more general way: \textit{let the arms parameterize each other}.  

Formally, we define another frame that takes the trajectory of the other arm as dynamic task parameters and represents the relative relationship as GMM as well. Different from the observation perspectives built with a fixed pose, the transformation matrices $\boldsymbol{A}_{c, t}, \boldsymbol{b}_{c, t}$ are dynamic that change with the motion of the other arm. The relative motion is described as $\boldsymbol{\zeta}_c=\boldsymbol{A}_{c, t}^{-1}\left(\boldsymbol{\xi}-\boldsymbol{b}_{c, t}\right)$ and represented by $\{\pi^{(k)},\boldsymbol{\mu}_c^{(k)},  \boldsymbol{\Sigma}_c^{(k)}\}_{k=1}^K$. For each arm $h$, the task-specific GMM obtained by PoE 
\begin{equation}
	\begin{small}
		\begin{aligned}
			\mathcal{N}\left(\boldsymbol{\hat{\nu}}^{(k)}, \boldsymbol{\hat{\Gamma}}^{(k)}\right) \propto \prod_{j=1}^P \mathcal{N}\left(\boldsymbol{\nu}_{j}^{(k)}, \boldsymbol{\Gamma}_{j}^{(k)}\right) \cdot \mathcal{N}\left(\boldsymbol{\nu}_{c}^{(k)}, \boldsymbol{\Gamma}_{c}^{(k)}\right)
		\end{aligned}
	\end{small}
\end{equation}
where $\boldsymbol{\nu}_{c}^{(k)}=\boldsymbol{A}_{c, t}\boldsymbol{\mu}_{c}^{(k)}+\boldsymbol{b}_{c, t}, \boldsymbol{\Gamma}_{j}^{(k)}=\boldsymbol{A}_{c, t}\boldsymbol{\Sigma}_{j}^{(k)}\boldsymbol{A}_{c, t}^\top$.

Such a relative parameterization entangles the representation of bimanual arms together, letting them consider each other by constructing time-varying mutual observing perspectives. This brings two useful functions:
\begin{itemize}  
	\item Generate the motion of one arm based on a given motion of the other one in a \textit{leader-follower} manner.
	\item Generate bimanual motions to adapt to new situations simultaneously in a \textit{synergistic} manner.
\end{itemize}

\begin{figure*}[ht]
	\centerline{\includegraphics[width=0.9\linewidth]{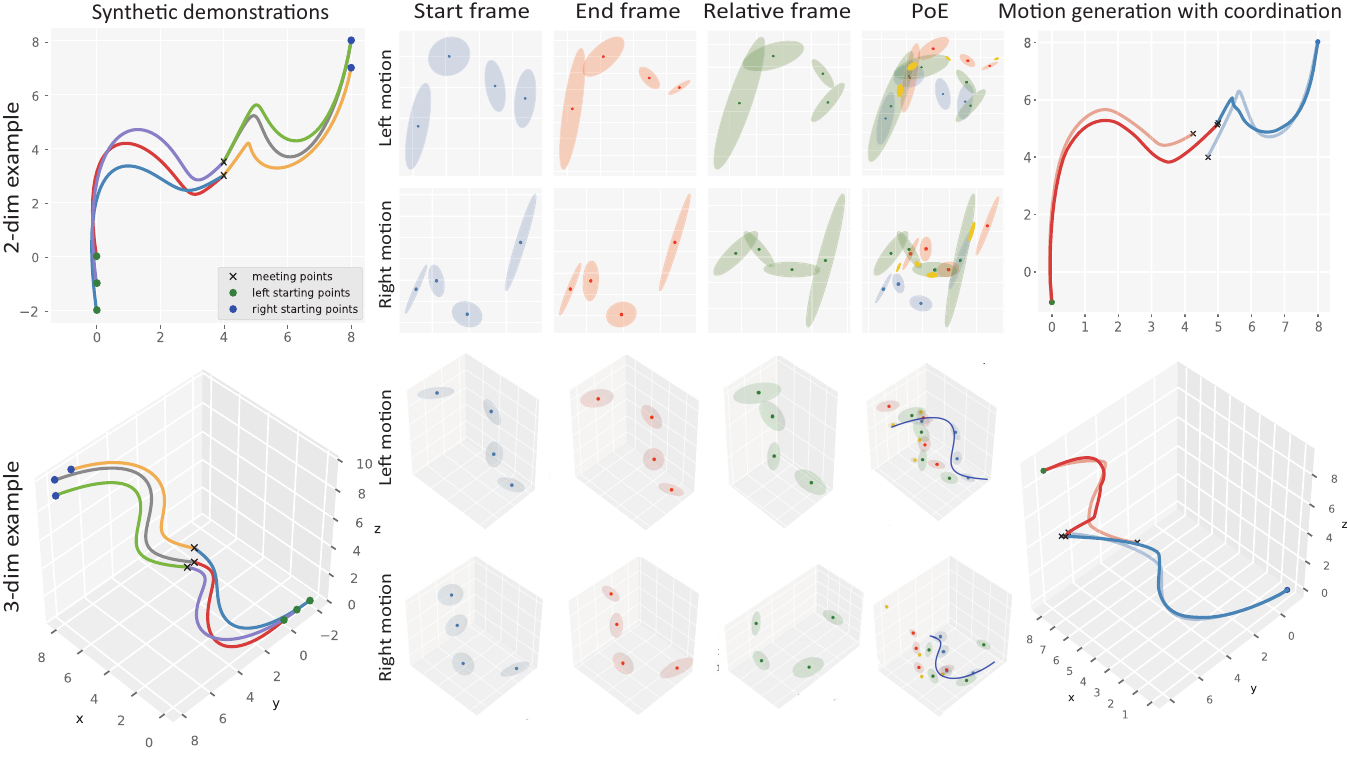}} 
	\caption{The upper and lower rows show the effect of the proposed relative parameterization method on bimanual coordination learning with 2-dim and 3-dim synthetic data, respectively. In each example, three synthetic bimanual motions are given as demonstrations and use the relative parameterization method to extract and construct a coordination relationship. The right column is the motion result under the new task parameters generated by the proposed method, which are compared with motion generation without coordination (trajectories in light colors).}
	\label{synthetic}
\end{figure*}
For instance, if the left arm motion $\boldsymbol{\xi}_l$ is pre-defined or adjusted to new situations by other methods like Dynamic Movement Primitive (DMP) in \cite{liu2022robot}, a corresponding right arm motion that considers the spatial-temporal coordination implicit in the demonstration can be generated by gaining the dynamic relative parameters $\boldsymbol{A}_{c, t}, \boldsymbol{b}_{c, t}$ from  $\boldsymbol{\xi}_l$. Then we can obtain a task-and-coordination-specific GMM of the right arm for further motion generation and control.

For generating bimanual motions synergistically, since the bimanual motions are unknown at the beginning, the relative parameterization cannot be established. Thus, we first use the product of GMMs in other reference systems to generate the independent motions of arms and then use these motions as the relative frame of the other arm to embed learned coordination iteratively.

\subsection{Control with Relative Parameterization}\label{IIC}
Coordination relationships can also be embedded when generating trajectories and corresponding control commands from GMM. Let the cost function of the vanilla LQT controller in Equ. \ref{LQT} be $\mathcal{C}_{vanilla}$. The composition cost function that takes coordination into account can then be written as 
\begin{equation}
	\begin{small}
		\begin{aligned}
			\mathcal{C}=\sum_h^H\mathcal{C}_{vanilla}^{h}  + \left(\boldsymbol{\nu}_c-\boldsymbol{x}_c\right)^\top \boldsymbol{Q}_c\left(\boldsymbol{\nu}_c-\boldsymbol{x}_c\right)
		\end{aligned}
	\end{small}
\end{equation}

By setting a similar linear system like Equ. \ref{linear_system}, the composition cost function can rewrite the cost function as
\begin{equation}
	\begin{small}
		\begin{aligned}
			\mathcal{C}=&\sum_h^H\Big[\left(\boldsymbol{\hat\nu}^h_u-\boldsymbol{u}^h\right)^\top \boldsymbol{\Omega}_u^h\left(\boldsymbol{\hat\nu}^h_u-\boldsymbol{u}^h\right)  + {\boldsymbol{u}^h}^\top \boldsymbol{R}^h \boldsymbol{u}^h\Big]\\
			&+\left(\boldsymbol{\nu}_{u, c}-\boldsymbol{u}_c\right)^\top \boldsymbol{\Omega}_{u, c}\left(\boldsymbol{\nu}_{u, c}-\boldsymbol{u}_c\right)
		\end{aligned}
	\end{small}
\end{equation}
where $\boldsymbol{\hat\nu}^h_u=\boldsymbol{S}_u^{-1}\left(\boldsymbol{\hat\nu}^h-\boldsymbol{S}_x x_1\right)$ and $\boldsymbol{\Omega}_u=\boldsymbol{S}_u^\top \boldsymbol{Q} \boldsymbol{S}_u$. $\boldsymbol{\hat\nu}_{u,c}$ and $\boldsymbol{\Omega}_{u,c}$ share the similar transformation.

Since there exists multivariate ($\boldsymbol{u}^h, \boldsymbol{u}_c$), we cannot directly change this sum of quadratic error terms into PoE. Thus, we set a unified vector $\boldsymbol{U} \in \mathbb{R}^{D T\times H}$ for representing the control command of the whole system, and a binary coordination matrix $\boldsymbol{C} \in \mathbb{R}^{D T \times D T H}, \boldsymbol{C}=[\boldsymbol{C}^1, \ldots, \boldsymbol{C}^H]$. For convenience, we set $[\boldsymbol{C}^h]=[\mathbf{0}, \ldots, \boldsymbol{C}^h, \ldots, \mathbf{0}]$, then we can continue to rewrite the cost function as
\begin{equation}
	\begin{small}
		\begin{aligned}
			\mathcal{C} =&\sum_h^H\Big[\left(\boldsymbol{\hat\nu}^h_u-[\boldsymbol{C}^h] \boldsymbol{U}\right)^\top \boldsymbol{\Omega}_u^h \left(\boldsymbol{\hat\nu}^h_u-[\boldsymbol{C}^h] \boldsymbol{U}\right) \\&+\boldsymbol{U}^\top[\boldsymbol{C}^h]^\top \boldsymbol{R}^h[\boldsymbol{C}^h] \boldsymbol{U}\Big]
			\\&+\left(\boldsymbol{\nu}_{u, c}-\boldsymbol{C} \boldsymbol{U}\right)^\top \boldsymbol{\Omega}_{u, c}\left(\boldsymbol{\nu}_{u, c}-\boldsymbol{C} \boldsymbol{U}\right)
		\end{aligned}
	\end{small}
\end{equation}

Set  $\boldsymbol{\Omega}_U^h=[\boldsymbol{C}^h]^\top \boldsymbol{\Omega}_u^h[\boldsymbol{C}^h]$, $\boldsymbol{R}_U^h=[\boldsymbol{C}^h]^\top \boldsymbol{R}^h[\boldsymbol{C}^h]$, $\boldsymbol{\hat\nu}^h_U=[\boldsymbol{C}^h]^{-1} \boldsymbol{\hat\nu}^h_u$, $\boldsymbol{\nu}_{U, c}=\boldsymbol{C}^{-1} \boldsymbol{\nu}_{u, c}$, the composition cost function is simplified as
\begin{equation}
	\begin{small}
		\begin{aligned}
			\mathcal{C} =&\sum_h^H\Big[\left(\boldsymbol{\hat\nu}^h_U-\boldsymbol{U}\right)^\top \boldsymbol{\Omega}_U^h \left(\boldsymbol{\hat\nu}^h_U-\boldsymbol{U}\right) +\boldsymbol{U}^\top \boldsymbol{R}^h_U \boldsymbol{U}\Big]
			\\&+\left(\boldsymbol{\nu}_{U, c}-\boldsymbol{U}\right)^\top \boldsymbol{\Omega}_{U, c}\left(\boldsymbol{\nu}_{U, c}-\boldsymbol{U}\right)
		\end{aligned}
	\end{small}
\end{equation}

Then we can finally change this sum of quadratic error terms into PoE
\begin{small}
	\begin{equation}
		\begin{aligned}
			\mathcal{N}&\left({\boldsymbol{\hat U}}, \boldsymbol{\hat{\Sigma}}_U\right) \propto \\&\prod_h^H\left[\mathcal{N}\left(0, {\boldsymbol{R}_U^h}^{-1}\right) \mathcal{N}\left(\boldsymbol{\hat\nu}^h_U, {\boldsymbol{\Omega}_U^h}^{-1}\right)\right] \mathcal{N}\left(\boldsymbol{\nu}_{U, c},{\boldsymbol{\Omega}_{U, c}}^{-1}\right)
		\end{aligned}
	\end{equation}
\end{small}

\begin{figure*}[ht]
	\centerline{\includegraphics[width=1\linewidth]{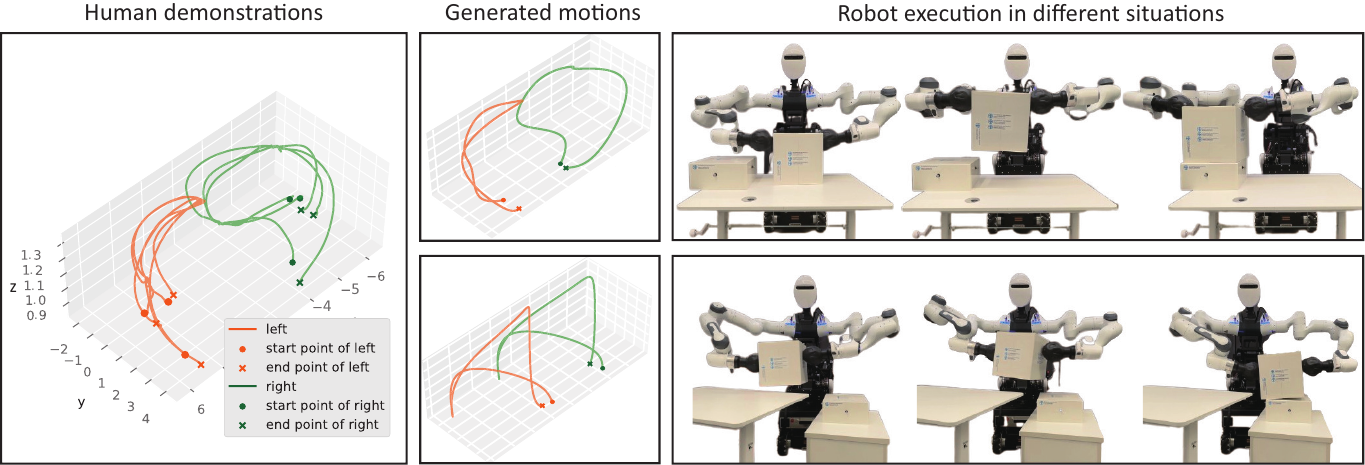}} 
	\caption{This figure shows the effect of the proposed relative parameterization method in the real bimanual robot manipulation with the example of palletizing. The task parameters in the robot execution process are different from those in human demonstrations, which reflects the generalizability of the proposed method. Meanwhile, the ability to maintain the \textit{synergistic} coordination in generalized motions is the core contribution of this work.}
	\label{real}
\end{figure*}

The result can be written as
\begin{equation}
	\begin{small}
		\begin{aligned}
			\boldsymbol{\hat{\Sigma}}_U& =\left(\sum_h^H\left[\boldsymbol{\Omega}_U^h+\boldsymbol{R}_U^h\right]+\boldsymbol{\Omega}_{U, c}\right)^{-1} \\
			\boldsymbol{\hat U}& =\boldsymbol{\hat{\Sigma}}_U\left(\sum_h^H \boldsymbol{\Omega}_U^h \boldsymbol{\hat\nu}^h_U+ \boldsymbol{\Omega}_{U, c}\boldsymbol{\nu}_{U, c}\right)
		\end{aligned}
	\end{small}
\end{equation}
By using the binary coordination matrix $\boldsymbol{C}$, we can extract the coordinated control commands and motions from $\boldsymbol{\hat U}$.

\subsection{Weighted Relative Parameterization}\label{IID}
A feasible variant of the above methods is to introduce weight coefficients $\sigma$ to adjust the influence of coordination relationship in representation and control. 

For the GMM representation,
\begin{equation}
	\begin{small}
		\begin{aligned}
			\mathcal{N}\left(\boldsymbol{\hat{\nu}}^{(k)}, \boldsymbol{\hat{\Gamma}}^{(k)}\right) \propto \prod_{j=1}^P \mathcal{N}\left(\boldsymbol{\nu}_{j}^{(k)}, \boldsymbol{\Gamma}_{j}^{(k)}\right) \cdot \left[\mathcal{N}\left(\boldsymbol{\nu}_{c}^{(k)}, \boldsymbol{\Gamma}_{c}^{(k)}\right)\right]^{\sigma}
		\end{aligned}
	\end{small}
\end{equation}

For the LQT controller,
\begin{equation} 
	\begin{small}
		\begin{aligned}
			\boldsymbol{\hat{\Sigma}}_U& =\left(\sum_h^H\left[\boldsymbol{\Omega}_U^h+\boldsymbol{R}_U^h\right]+{\sigma}\cdot\boldsymbol{\Omega}_{U, c}\right)^{-1} \\
			\boldsymbol{\hat U}& =\boldsymbol{\hat{\Sigma}}_U\left(\sum_h^H \boldsymbol{\Omega}_U^h \boldsymbol{\hat\nu}^h_U+ {\sigma}\cdot\boldsymbol{\Omega}_{U, c}\boldsymbol{\nu}_{U, c}\right)
		\end{aligned}
	\end{small}
\end{equation}


\section{Experiments}

\subsection{Setup}
The effectiveness of the proposed method is illustrated by learning through both synthetic motions and real demonstration motions. Some pre-designed coordinated motions can show the coordination explicitly, which is meant to demonstrate the performance of the method.

\textbf{Synthetic motions:} The synthetic motions were created via B\'ezier curves, where bimanual arms depart from a distance and meet at the same point. This kind of motion often occurs in some daily  activities that require both arms to grasp, carry or pick up something simultaneously. We provide both two-dimensional and three-dimensional data to show the dimension scalability, as shown in Fig. \ref{synthetic}. 

\textbf{Real demonstration motions:} We also provide demonstrations of two real tasks to show the effect in bimanual robot manipulation. The palletizing example shown in Fig. \ref{real} represents a class of \textit{synergistic} coordinated motions and tasks, while the pouring example shown in \ref{framework} is a typical bimanual coordination task in the \textit{leader-follower} manner.

\subsection{Demonstration collection}
The human demonstration data was collected via Optitrack. The demonstrator attached two groups of markers on his hands for detection by Optitrack. Each group of markers contains four individual markers, which are required to determine the pose of each arm. These four markers will be detected via six Optitrack cameras to record two end-effector trajectories with both position and orientation. We chose the poses from the centers of each marker group to reproduce human bimanual demonstration motions. In addition, the box and the two cups each have a set of four markers for recording object motions. The raw data were pre-processed by our open-source toolbox \cite{junjia_liu_2023_8084510} to extract the valuable information and separate it into multiple demonstrations visually. Each demonstration will have seven pose values for each marker group.

\subsection{Coordination learning performance analysis}
The goal of synthetic motions is that bimanual arms should meet in the same pose, whether in 2-dim or 3-dim. As shown in the left column of Fig. \ref{synthetic}, we provide three bimanual motions as demonstrations for each synthetic example. These motions start and end from different positions but move in a similar style. The middle column, with multiple small figures, shows the process of using the proposed relative parameterization method. We use three observation frames to parameterize the motions of each arm, from the start points, endpoints, and a dynamic relative observation frame depending on the other arm. We can extract and construct coordination relationships from this parameterization from demonstration data. The parameterized coordination is then used in motion generation and control in new situations with different task parameters. Keeping the same coordination relationship in these generalized motions is required to achieve some specific bimanual tasks. Thus, the generalized motion generation results are shown in the right column of Fig. \ref{synthetic}. In the 2-dim example, bimanual motions are required to meet at a new position, $(5, 5)$. In the 3-dim example, this new meeting point is set to $(5, 8, 5)$. The generated motions with learned coordination are shown in red and blue, while we also provide a comparison with generated motions without coordination (in light red and blue). By comparison, we find that just regarding bimanual arms as a simple combination of two single arms is insufficient for bimanual tasks. It is necessary to parameterize coordination relationship no matter in a \textit{leader-follower} or \textit{synergistic} manner; this is the key to achieving bimanual tasks mostly.

\subsection{Real robot experiment}
We adopt the self-designed humanoid CURI robot for real robot experiments to perform the bimanual motions. Since this work focuses on learning and generalizing coordinated motion, task parameters such as start and end points and object poses are obtained through the Optitrack system. As shown in Fig. \ref{real}, we paste four markers on the box to be transported and the box as a destination to facilitate obtaining their poses in the world coordinate system. Meanwhile, four fixed connected markers are also on the back of the CURI robot. The coordinated human hand motions are learned by relative parameterization. Then we use this parameterized coordination model to generate motions that adapt to new object poses and destinations. It is worth mentioning that, unlike the observation frames used in the synthetic data, we set five observation frames to transport this palletizing task, namely from start points, end points, center poses of the transport box, and the center pose of the destination box. This allows the robot to move from an initial pose with its arms outstretched to the sides of the box, carry the box and place it in the target position, and then release the box. Besides, the result of the pouring example can be found in Fig. \ref{framework}. A self-designed impedance controller supports the execution of the CURI robot, and the trajectories are converted to joint space commands via its inverse kinematics model.


\section{Discussion}
This work still has some limitations. First, the proposed relative parameterization method is only applied to trajectories in Cartesian space without considering joint space coordination. Learning joint-space bimanual coordination or even whole-body coordination from human demonstrations remains an open problem. Some previous work can be found in \cite{silverio2018bimanual}. Besides, the method based on the Gaussian mixture model will take a certain amount of time when processing high-frequency sampling demonstration data, which might affect the actual real-time usage. Some improvements using Tensor instead of large sparse matrices can be found in \cite{shetty2021ergodic}.

\section{Conclusion}
In this work, we propose a method for parameterizing coordination in bimanual tasks by probabilistic relative motion relationship of bimanual arms from human demonstration and guiding the robot motion generation in new situations. By embedding relative motion relationship, bimanual motions can be generated in a \textit{leader-follower} manner and also \textit{synergistic} manner. We provide a detailed formulation derivation process and demonstrate the effectiveness of the proposed method in coordination learning with some synthetic data with prominent coordination characteristics. We also deploy the method on a real humanoid robot to perform coordination motions to show its generalization in new situations. We believe that this easy-to-use bimanual LfD method can be used as a robust demonstration data augmentation method for training robot large manipulation model \cite{liu2023softgpt}, and we will do research to show this potential in the future.

\bibliographystyle{ieeetr}
\bibliography{ref_rp}

\end{document}